\documentclass{article}

\usepackage{array}
\usepackage{arxiv}
\usepackage{array}
\usepackage[utf8]{inputenc} % allow utf-8 input
\usepackage[T1]{fontenc}    % use 8-bit T1 fonts
\usepackage{hyperref}       % hyperlinks
\usepackage{url}            % simple URL typesetting
\usepackage{booktabs}       % professional-quality tables
\usepackage{amsfonts}       % blackboard math symbols
\usepackage{nicefrac}       % compact symbols for 1/2, etc.
\usepackage{microtype}      % microtypography
\usepackage{lipsum}
\usepackage{graphicx}
\usepackage{float}
\usepackage{algorithm}
\usepackage{algorithmic}
\usepackage{amsmath}
\graphicspath{ {./images/} }

\title{Visual Error Patterns in Multi-Modal AI: \\ A Statistical Approach}

\author{
 Ching-Yi Wang \\
 \\
  Department of Psychology, University of California, Los Angeles, Los Angeles, CA 90095\\
  Department of Mathematics, University of California, Los Angeles, Los Angeles, CA 90095\\
  \\
  \texttt{ian0504@ucla.edu} 
}

\begin{document}
\maketitle
\begin{abstract}
Multi-modal large language models (MLLMs), such as GPT-4o, excel at integrating text and visual data but face systematic challenges when interpreting ambiguous or incomplete visual stimuli. This study leverages statistical modeling to analyze the factors driving these errors, using a dataset of geometric stimuli characterized by features like ‘3D,’ ‘rotation,’ and ‘missing face/side.’ We applied parametric methods, non-parametric methods, and ensemble techniques to predict classification errors, with the non-linear gradient boosting model achieving the highest performance (AUC = 0.85) during cross-validation. Feature importance analysis highlighted difficulties in depth perception and reconstructing incomplete structures as key contributors to misclassification. These findings demonstrate the effectiveness of statistical approaches for uncovering limitations in MLLMs and offer actionable insights for enhancing model architectures by integrating contextual reasoning mechanisms.
\end{abstract}

% keywords can be removed

\section{Introduction}
\maketitle
Artificial Intelligence (AI) has achieved transformative success across a wide range of domains, revolutionizing fields such as healthcare, education, and human-computer interaction [1]. However, the mechanisms driving AI's performance often remain opaque, particularly in the context of large language models (LLMs), which have advanced at an unprecedented pace in recent years. Multi-modal large language models (MLLMs) like GPT-4o exemplify this evolution, integrating text, audio, and visual inputs to enable interaction across diverse domains. Despite their remarkable capabilities, these models remain largely "black boxes," offering limited insight into how they process multi-modal information internally. This lack of transparency poses significant risks, including systematic biases, flawed associations, and unintended behaviors, which demand urgent attention [7]. Understanding the decision-making processes of MLLMs is not only beneficial but essential for mitigating these risks and ensuring their safe deployment in critical applications.

GPT-4o was chosen as the focus of this study for its advanced multi-modal capabilities, which allow simultaneous processing of textual and visual information [4]. These capabilities make it an ideal model for investigating the parallels and distinctions between machine-driven and human-driven visual perception. While GPT-4o excels in tasks involving structured and complete data, its reliance on bottom-up processing—a feature-by-feature analysis of sensory inputs—renders it less effective at interpreting complex or ambiguous stimuli. This limitation contrasts sharply with human vision, which is remarkably adept at resolving ambiguity and reconstructing incomplete information through high-level cognitive processes [11, 13].

Human vision uses top-down processing, a mechanism that integrates prior knowledge, experiences, and expectations to interpret sensory inputs, enabling rapid and accurate perception even in the face of uncertainty. Gestalt principles, such as the Law of Closure and Amodal Completion, exemplify this ability, allowing humans to infer missing elements of objects. In contrast, bottom-up processing, the primary mechanism used by GPT-4o, relies solely on raw sensory data, lacking the contextual inference and mental frameworks that characterize human cognition [8, 9]. This fundamental distinction highlights the challenges faced by MLLMs in handling visual classification tasks and underscores the importance of investigating cognitive-inspired approaches to improve these models.

\begin{itemize}
    \item[1] \textbf{Bottom-Up Limitation Hypothesis}: Errors in MLLMs arise primarily from their dependence on raw sensory data, which limits their ability to integrate contextual inference or prior knowledge.
    \item[2] \textbf{Feature-Specific Error Hypothesis}: Errors are driven more by the inherent complexity of specific visual features, such as 3D structures, rotations, or missing faces, rather than the absence of contextual reasoning [2].

\end{itemize}

This study aims to explore whether MLLMs like GPT-4o encounter specific challenges in visual classification due to their reliance on bottom-up processing. Two hypotheses guide our investigation:

The Bottom-Up Limitation Hypothesis: Errors in MLLMs arise primarily from their dependence on raw sensory data, which limits their ability to integrate contextual inference or prior knowledge.
The Feature-Specific Error Hypothesis: Errors are driven more by the inherent complexity of specific visual features, such as 3D structures, rotations, or missing faces, rather than the absence of contextual reasoning [5, 6].

To evaluate these hypotheses, we designed a dataset of 54 3D and 21 2D geometric stimuli, featuring shapes like cubes, prisms, pentagons, and hexagons. These stimuli were characterized by attributes such as missing faces, rotations, and stacked elements, designed to test the model's ability to interpret visual complexity. We assessed GPT-4o’s performance using multi-regression analysis to quantify how individual visual features contributed to error rates. Our analysis focused on the model's capacity to distinguish between geometrically similar shapes, reconstruct incomplete structures, and resolve ambiguities in orientation and arrangement. The findings offer critical insights into the limitations of MLLMs in visual classification and potential improvement inspired by human cognitive processes [3].

\section{METHODS}
\subsection{Dataset}
The dataset used in this study comprised 75 carefully curated visual stimuli designed to evaluate the performance of multi-modal large language models (MLLMs) in visual classification tasks. Each stimulus was characterized by a combination of geometric features, including '3D,' 'Circle,' 'Pentagon,' 'Hexagon,' 'Cube,' and 'Triangle,' alongside spatial attributes such as arrangement patterns, repetition, combination, rotation, and missing faces. These features were selected to represent diverse structural and spatial challenges, emphasizing complex arrangements and incomplete structures to simulate real-world scenarios where stimuli are often ambiguous or incomplete. For example, shapes were presented with varying orientations, stacked or overlapped in systematic patterns, or featured missing surfaces to test the model’s ability to interpret depth cues, recognize patterns, and reconstruct partial information. This dataset was engineered to address two objectives: (1) to introduce visual complexity by incorporating ambiguous and challenging stimuli, and (2) to ensure feature diversity by including a wide spectrum of geometric and spatial attributes. By focusing on these aspects, the dataset provides a unique testing ground for probing the strengths and limitations of MLLMs, such as GPT-4o, offering insights into their capacity to process complex visual inputs and handle edge cases that are often overlooked in traditional datasets.
\\
\newpage
\subsection{Feature Classification:}
GPT-4o’s outputs were evaluated to classify errors based on visual features. Errors were categorized by:
% Table
% Table
\begin{table}[h!]
\centering
\renewcommand{\arraystretch}{1.5} % Adjust row height
\setlength{\tabcolsep}{8pt} % Adjust column separation
\caption{Error Rates by Stimulus Type and Feature}
\label{tab:error_rates}
\begin{tabular}{|>{\raggedright\arraybackslash}m{3.5cm}|>{\raggedright\arraybackslash}m{10cm}|}
\hline
\textbf{Error Type} & \textbf{Description of Error Rates} \\ \hline
Shape Type & High error rate in 3D classification (32.7\%), while 2D error rate remains low ($<$5\%). \\ \hline
Orientation & Minimal error in 3D tasks ($<$5\%), but moderate error (14.3\%) in 2D tasks. \\ \hline
Symmetry & Low error rates in both 3D and 2D classifications ($<$5\%). \\ \hline
Depth Cues & Minimal error in both 3D and 2D tasks ($<$5\%). \\ \hline
Missing Face & Significant error in 3D (63.2\%) and moderate error in 2D tasks (25\%). \\ \hline
Arrangement & Low error in 3D tasks ($<$5\%) but moderate error ($\sim$10\%) in 2D tasks. \\ \hline
Overall Structure & Moderate error in both 3D and 2D tasks ($\sim$10\%). \\ \hline
\end{tabular}
\end{table}

\subsection{Prompt Design:}
The following prompt was used to evaluate GPT-4o's ability to analyze visual stimuli:
\textit{\\
"Please analyze the uploaded image and provide a detailed, text-based description of the features you can extract. Specifically, identify key attributes such as shapes, orientation, depth cues, symmetry, arrangement, rotation, and any other noticeable elements in the image. If the image contains two different shapes, describe it as a 'combination.' If the same shape is repeated multiple times, describe it as 'repetition.' For any other configurations, provide a description of the observed structure."}

\subsection{Statistical Models}
After classifying the errors based on the all input images, we aimed to investigate the factors contributing to visual classification errors and perform feature importance analysis using supervised learning techniques. To achieve these objectives, we applied four statistical and machine learning models based on their unique strengths in capturing both linear and non-linear relationships [12].

\begin{itemize}
    \item [1] Logistic Regression: Baseline model to predict classification errors and interpret linear relationships between features and outcomes.
    
    \item [2] Ridge Logistic Regression: Using L2regularization to reduce overfitting, stabilizing predictions in the presence of multicollinearity.
    
    \item[3] Random Forest: Implemented with hyperparameter tuning (e.g., number of trees, maximum depth, minimum samples split) to optimize performance and prevent overfitting while leveraging its robust handling of non-linear relationships and feature interactions.
    
    \item[4] Gradient Boosting (XGBoost): Includes hyperparameter tuning (e.g., learning rate, maximum depth, number of estimators, subsample ratio) to achieve optimal performance and ensure efficient model convergence, while identifying influential features. The hyperparameters were explored across the following ranges: learning rate (0.01–0.3) to control weight updates and prevent overfitting, maximum depth (3–10) to balance model complexity and generalizability, number of estimators (50–200) to control the number of boosting iterations, and subsample ratio (0.5–1.0) to reduce overfitting by using random subsets of training data. 
    
\end{itemize}
%%%%%%%%%%%%%%%%%%%%%%%%%%%%%%%%%%%%%%%%%%%%%%%%%%
\newpage
\begin{algorithm}
\caption{Hyperparameter Tuning for XGBoost}
\label{alg:xgboost_tuning}
\begin{algorithmic}[1]
\REQUIRE Training data $(X_{\text{train}}, y_{\text{train}})$, parameter grid $G$, evaluation metric $\mathcal{M}$ (e.g., AUC)
\ENSURE Optimal hyperparameters $\Theta^*$

\STATE Initialize parameter grid $G$ with ranges for:
    \begin{itemize}
        \item Learning rate ($\eta$)
        \item Maximum tree depth ($d$)
        \item Number of estimators ($n$)
        \item Subsample ratio ($\text{subsample}$, if applicable)
        \item Regularization parameters ($\lambda, \alpha$)
    \end{itemize}
\STATE Initialize an empty dictionary $\mathcal{S}$ to store scores for each combination of parameters
\FOR{each combination of parameters $\Theta \in G$}
    \STATE Train the XGBoost model on $(X_{\text{train}}, y_{\text{train}})$ using hyperparameters $\Theta$
    \STATE Evaluate the model on the validation set using metric $\mathcal{M}$ and record the score $\mathcal{M}_\Theta$
    \STATE Store $(\Theta, \mathcal{M}_\Theta)$ in $\mathcal{S}$
\ENDFOR
\STATE Select $\Theta^* = \arg\max_{\Theta \in G} \mathcal{M}_\Theta$
\RETURN $\Theta^*$
\end{algorithmic}
\end{algorithm}

\subsection{Evaluation Metrics}
To evaluate the performance and robustness of the models, the following metrics were employed:
\begin{itemize}
    \item Area Under the Curve (AUC): AUC measures the model's ability to discriminate between different classes, providing a comprehensive evaluation of its predictive performance. A higher AUC indicates better performance in correctly distinguishing between error and non-error cases.

    \item Receiver Operating Characteristic (ROC) Curve: The ROC curve illustrates the trade-off between true positive rate (TPR) and false positive rate (FPR) across various classification thresholds. It provides a visual representation of model performance, with the curve's proximity to the top-left corner indicating high discriminative ability.

    \item Cross-Validation (5-fold): A 5-fold cross-validation strategy was used to assess the robustness and generalizability of the models. The dataset was partitioned into five subsets, with the model trained on four subsets and validated on the fifth, repeated for all combinations. This ensured consistent evaluation across different splits and minimized overfitting.
\end{itemize}

\subsection{Example Stimuli}
\begin{figure}[H]
    \centering
    \includegraphics[width=6cm]{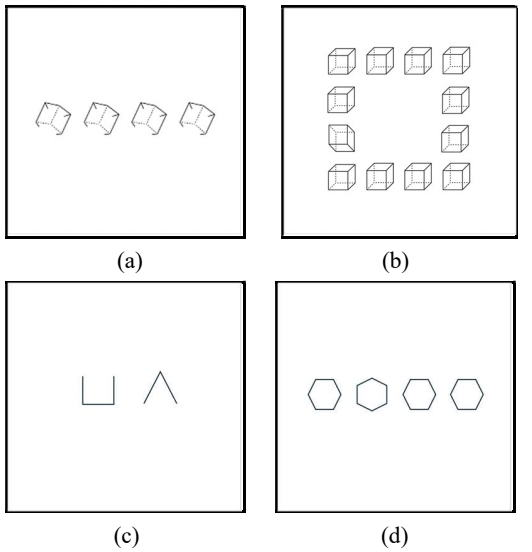}
    \caption{Example visual stimuli from the dataset. (a) A 3D pentagonal prism with a missing face. (b) A 3D arrangement of cubes in a square pattern, with one cube rotated upside down relative to the others. (c) A combination of a 2D square and a triangle missing one side. (d) Repetition of 2D hexagons arranged linearly, with one hexagon in the center rotated.}
    \label{fig:stimuli}
\end{figure}

\section{RESULTS:}
We evaluated GPT-4o's performance on a dataset comprising 75 geometric stimuli, including 54 3D and 21 2D shapes. These stimuli were systematically designed to include attributes such as missing faces, rotations, and stacked elements, challenging the model's ability to process structural complexities and ambiguities. Each stimulus was binary-coded to indicate whether the model classified it correctly, enabling quantitative analysis of error rates. A multi-regression analysis was conducted to examine the contributions of specific visual features—such as symmetry, orientation, and depth—to classification errors, integrating both qualitative insights and quantitative metrics. This approach provided a comprehensive evaluation of GPT-4o's strengths and limitations in visual feature classification.

\subsection{Error Evaluation}
For 3D stimuli, GPT-4o demonstrated substantial difficulty in distinguishing geometrically similar shapes. For example, pentagonal and hexagonal prisms were frequently misclassified, resulting in a 32.7\% error rate for shape type. Furthermore, the model struggled significantly with incomplete structures, exhibiting a 63\% error rate when identifying shapes with missing faces. These findings underscore the model's limitations in processing partial or ambiguous inputs.

In contrast, GPT-4o performed exceptionally well in tasks requiring depth perception, achieving a low overall error rate of 3.7\% for 3D structure identification. This suggests that the model excels in recognizing spatial relationships and depth cues, particularly in well-defined and complete structures.

For 2D stimuli, the model's most frequent classification error involved orientation, with a 14.3\% error rate. This limitation likely stems from the absence of depth cues, which are critical for resolving ambiguities in object rotation and alignment. While GPT-4o achieved high accuracy for simpler features, such as symmetry and shape type, its performance diminished as the visual complexity of stimuli increased. Overlapping shapes and rotated configurations, in particular, exacerbated classification errors.

\begin{table}[h!]
\centering
\renewcommand{\arraystretch}{1.5} % Adjust row height
\setlength{\tabcolsep}{8pt} % Adjust column separation
\caption{Error Rates by Stimulus Type and Feature}
\label{tab:error_rates}
\begin{tabular}{|>{\raggedright\arraybackslash}m{3.5cm}|>{\raggedright\arraybackslash}m{3.5cm}|>{\raggedright\arraybackslash}m{3.5cm}|}
\hline
\textbf{Error Type} & \textbf{3D Error Rate (\%)} & \textbf{2D Error Rate (\%)} \\ \hline
Shape Type & 32.7\% & Low ($<$5\%) \\ \hline
Orientation & Low ($<$5\%) & 14.3\% \\ \hline
Symmetry & Low ($<$5\%) & Low ($<$5\%) \\ \hline
Depth Cues & Low ($<$5\%) & Low ($<$5\%) \\ \hline
Missing Face & 63.2\% & 25\% \\ \hline
Arrangement & Low ($<$5\%) & Moderate ($\sim$10\%) \\ \hline
Overall Structure & Moderate ($\sim$10\%) & Moderate ($\sim$10\%) \\ \hline
\end{tabular}
\end{table}

An in-depth analysis of error rates revealed that GPT-4o encountered the most significant challenges with 3D stimuli, particularly with Missing Faces and Shape Type. The model exhibited a high error rate of 63.2\% for shapes with missing faces, reflecting its difficulty in reconstructing incomplete structures. Similarly, a 32.7\% error rate for Shape Type indicates substantial confusion in distinguishing geometrically similar shapes, such as pentagonal and hexagonal prisms. For 2D stimuli, the model's performance was most affected by Orientation, with an error rate of 14.3\%, suggesting challenges in interpreting rotated or misaligned shapes in the absence of depth cues. Additionally, a 25\% error rate for Missing Faces in 2D stimuli further underscores GPT-4o's limitations in processing incomplete and ambiguous visual features.\\
\newpage
% Table
\begin{table}[h!]
\centering
\renewcommand{\arraystretch}{1.5} % Adjust row height
\setlength{\tabcolsep}{8pt} % Adjust column separation
\caption{Cross-Validation Results for Different Models}
\label{tab:CV_test}
\begin{tabular}{|>{\raggedright\arraybackslash}m{3.5cm}|>{\raggedright\arraybackslash}m{3.5cm}|>{\raggedright\arraybackslash}m{3.5cm}|}
\hline
\textbf{Model} & \textbf{Mean AUC} & \textbf{Standard Deviation} \\ \hline
Logistic Regression & 0.79 & 0.03 \\ \hline
Ridge Logistic Regression & 0.79 & 0.03 \\ \hline
Random Forest & 0.80 & 0.09 \\ \hline
XGBoost & 0.85 & 0.02 \\ \hline
\end{tabular}
\end{table}

\subsection{MODEL RESULTS}
We evaluated the predictive performance of four statistical models—Logistic Regression, Ridge Logistic Regression, Random Forest, and Gradient Boosting (XGBoost)—to predict the occurrence of missing face errors. This task challenges the ability to model top-down inference. Logistic Regression served as a baseline, modeling linear relationships between input features and the occurrence of errors, while Ridge Logistic Regression incorporated $L_2$-regularization to control multicollinearity and stabilize predictions. Random Forest, an ensemble method utilizing bagging, captured non-linear relationships and feature interactions through randomized decision trees, offering robustness to noise and overfitting. XGBoost, a gradient boosting algorithm, iteratively optimized a differentiable loss function to minimize prediction errors, applying its ability to capture complex, non-linear feature interactions.

Each model was evaluated using a 5-fold cross-validation strategy, splitting the dataset into training and validation subsets to assess generalizability. The Area Under the Curve (AUC) was used as the primary evaluation metric to measure the model's discriminative ability in predicting whether missing face errors occurred. Missing face errors are particularly challenging because they require simulating top-down inference akin to human cognition, where prior knowledge and contextual cues are integrated to infer missing structures. Unlike simple classification tasks, accurately predicting missing faces relies on a model's ability to combine geometric and contextual information, extending beyond traditional bottom-up processing. This study aimed to assess the ability of these statistical models to meet this challenge and uncover key insights into their strengths and limitations in handling such inferential tasks.
\\
\begin{figure}[H]
    \centering
    \begin{minipage}{0.48\textwidth}
        \centering
        \includegraphics[width=5cm]{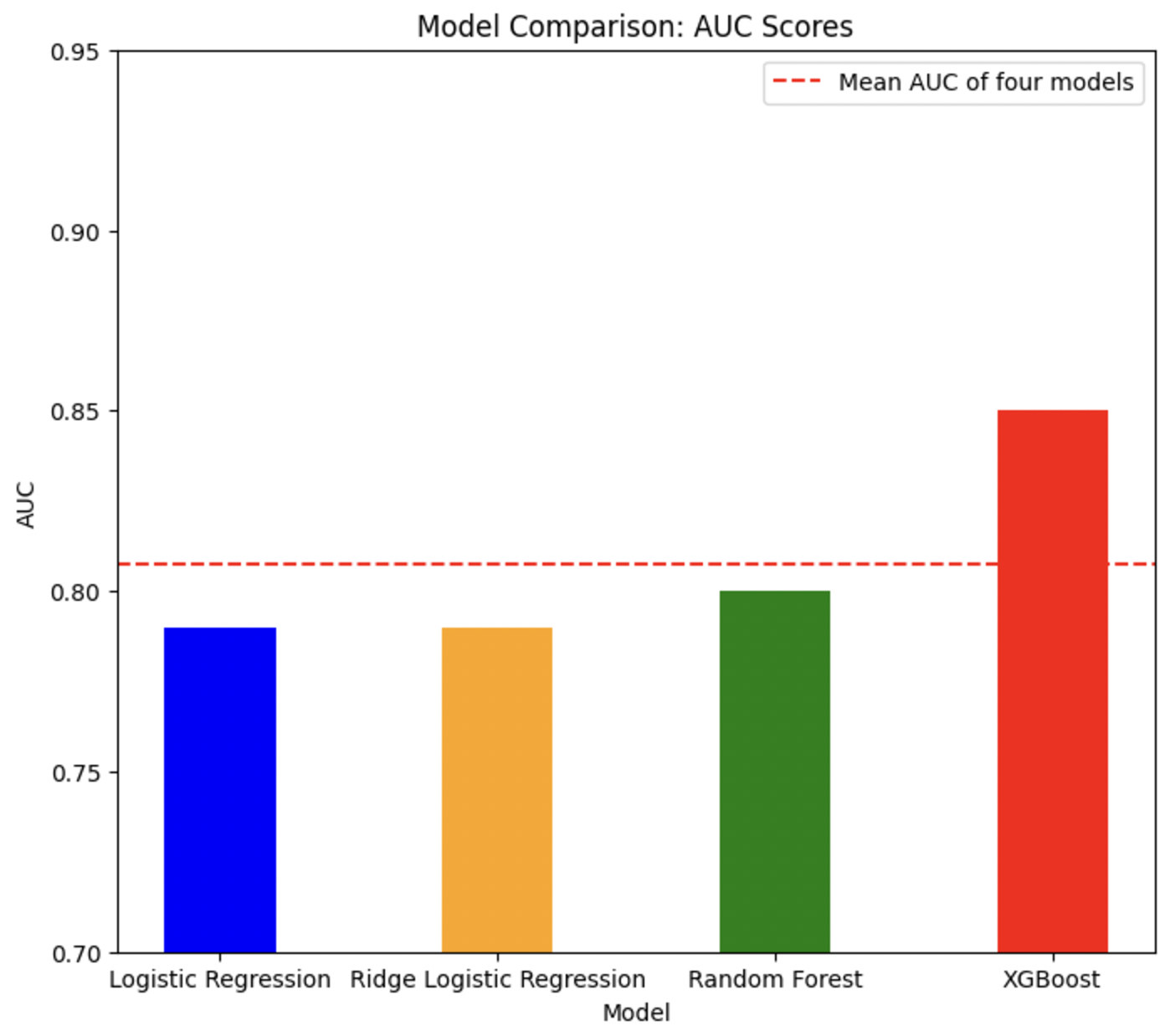}
        \caption{Bar plot of mean AUC scores for different models.}
        \label{fig:bar_plot}
    \end{minipage}
    \hfill
    \begin{minipage}{0.48\textwidth}
        \centering
        \includegraphics[width=5.5cm]{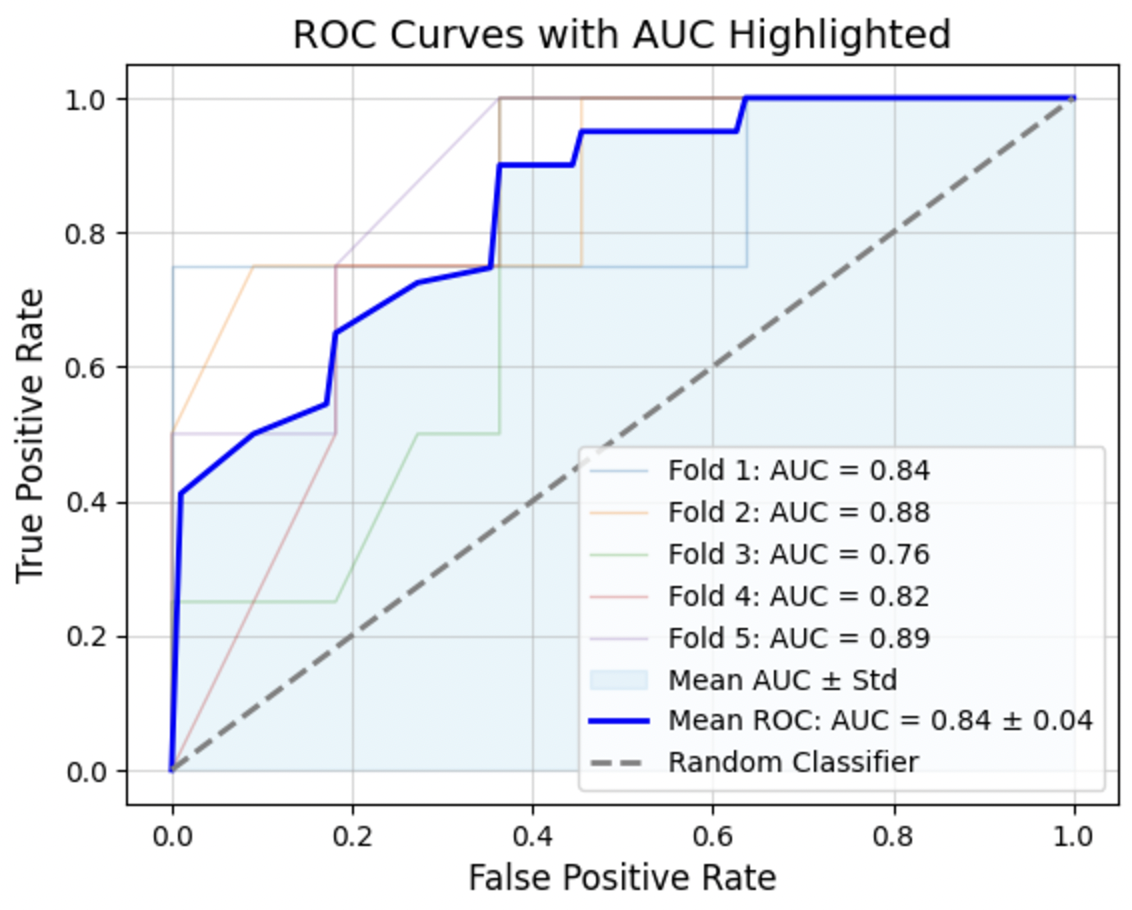}
        \caption{ROC curves with AUC highlighted for XGBoost during 5-fold cross-validation.}
        \label{fig:roc_curves}
    \end{minipage}
\end{figure}

\subsection{Feature Importance Analysis}
The feature importance analysis reveals the relative contribution of each input feature to the model's decision-making process, offering insights into the predictors that most significantly influence performance. Using XGBoost, we quantified the impact of features in predicting classification outcomes for the 'Missing Face' error. This analysis highlights the model's reliance on structural and geometric features, such as '3D' (33.24\%) and 'Pentagon' (33.61\%), which are critical in addressing the challenges of handling missing faces and complex arrangements.

Notably, features like '3D' highlights the model's difficulty with depth-related tasks, such as distinguishing between similar 3D shapes or interpreting incomplete structures. Missing faces in 3D objects disrupt spatial relationships, often leading to incorrect classifications. Humans, in contrast, resolve such ambiguities using top-down cognitive processes like Amodal Completion, which enables mental reconstruction of incomplete forms.

Lesser features, such as 'Triangle' (0.06\%) and 'Rotation' (2.14\%), have minimal influence, likely due to their simplicity or infrequency in the dataset. These findings suggest that enhancing the model’s ability to handle complex visual features and incomplete spatial information could significantly reduce classification errors. Future improvements should focus on minimizing ambiguities and integrating mechanisms that emulate human-like reasoning to interpret missing or complex visual stimuli effectively.

\begin{figure}[H]
    \centering
    \includegraphics[width=7cm]{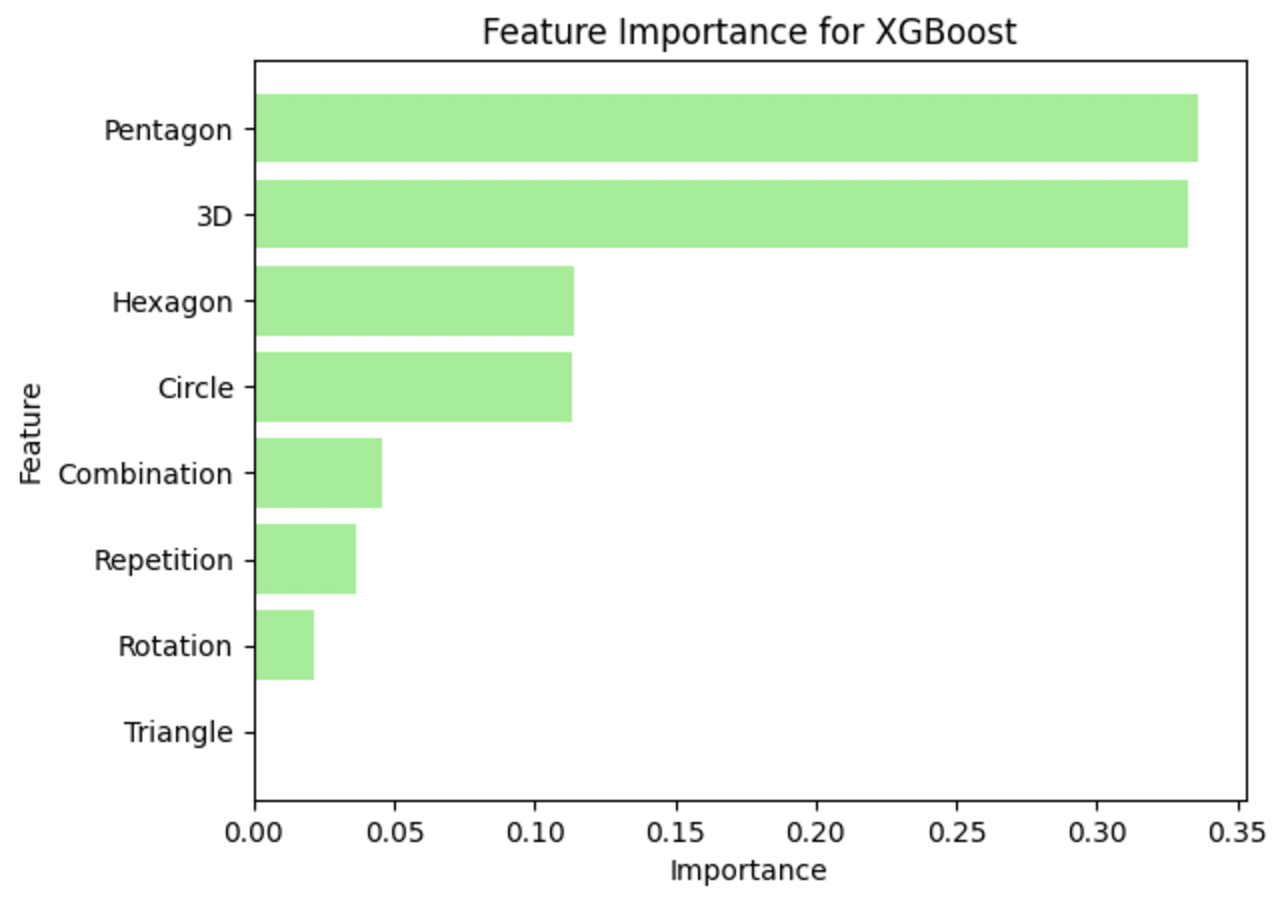}
    \caption{Bar plot of Feature Importance for XGBoost Model.}
    \label{fig:barplot for feature}
\end{figure}

\section{Discussion}
The purpose of the present study was to examine the performance of multi-modal large language models (MLLMs), specifically GPT-4o, in visual classification tasks involving geometric stimuli. Our primary goal was to understand the factors contributing to classification errors, particularly for features like ‘3D’ structures and ‘missing faces,’ and to explore whether the dependence on bottom-up processing limits the model’s ability to handle complex visual stimuli. We hypothesized that (a) the model would struggle with tasks requiring top-down inference, such as reconstructing missing elements or resolving ambiguities, or (b) the model would perform similarly to human perception when handling simpler geometric features by using its advanced processing capabilities.

Our results showed that statistical models, particularly XGBoost, effectively predicted classification errors, achieving the highest average AUC (0.85, SD = 0.02) compared to both linear models (Logistic Regression, Ridge Logistic Regression) and another non-linear model (Random Forest). The analysis of the feature importance identified ‘3D,’ ‘Circle,’ and ‘Pentagon’ as the most influential predictors of top-down errors, highlighting the challenges caused by geometric and structural complexity. These findings support our hypothesis that MLLMs such as GPT-4o rely heavily on bottom-up processing, which limits their ability to infer missing elements or resolve ambiguous visual stimuli. For instance, errors involving ‘3D’ features often came from the model’s difficulty in interpreting depth cues or incomplete structures, such as missing faces, where humans tend to perform better due to top-down processing mechanisms like Amodal completion [8]. Our findings highlight the limitations of GPT-4o’s feature-by-feature processing approach, particularly in tasks requiring reconstruction of incomplete structures, which aligns with broader challenges in multi-modal LLM development [10].

From the analysis, it is evident that GPT-4o struggles with visual ambiguity and incomplete information, failing to infer missing faces or reconstruct objects with disrupted symmetry.  This limitation is particularly concerning for features like ‘3D’ and ‘rotation,’ where the absence of contextual reasoning leads to misclassifications. While the model excels with simpler features like circles or cubes, its errors with complex stimuli emphasize a significant gap compared to human top-down processing.

Our observations reveal that GPT-4o is dependent on raw data which limits its robustness in complex scenarios. Errors with features like ‘rotation’ and ‘repetition’ often occurred in unusual orientations or overlapping patterns. This emphasizes the need for human-like strategies, such as top-down processing into MLLMs to enhance their ability to process incomplete or ambiguous visual data.

Our findings are limited by the scope of the dataset, which consisted of 75 geometric stimuli. While the stimuli were designed to capture a diverse range of structural and spatial attributes, a larger and more varied dataset could provide a deeper understanding of the model’s performance across different types of visual complexity. In addition, the study focused primarily on geometric features, such as shape type and depth cues, without exploring other potentially relevant factors like texture or shading. Future studies could expand the feature set to include such attributes, as well as test the model’s performance on more dynamic or real-world stimuli.

Despite these limitations, this study highlights important implications for AI development and evaluation, emphasizing the role of feature importance analysis in identifying error predictors and improving model robustness. Integrating human-like top-down inference could enhance MLLMs’ ability to process ambiguous and incomplete visual data.
This study highlights the importance of statistical modeling in analyzing and predicting errors in MLLMs. By uncovering the relationship between geometric and structural features and misclassifications, we offer a pathway to enhancing model robustness. Addressing these limitations will be critical as MLLMs find broader applications in real-world scenarios that demand reliable visual perception and decision-making.

\section{Conclusion}

This study used statistical modeling to evaluate the performance of GPT-4o in visual classification tasks involving geometric stimuli, focusing on misclassification predictors like '3D' structures and 'missing faces.' XGBoost achieved the highest predictive accuracy with an AUC of 0.85 (SD = 0.02) which outperforming both linear models (Logistic Regression, Ridge Logistic Regression) and Random Forest. Feature importance analysis identified '3D,' 'Circle,' and 'Pentagon' as key contributors to errors, demonstrating a significant relationship between geometric complexity and classification performance. These results highlight GPT-4o’s reliance on bottom-up processing, which limits its ability to handle ambiguous or incomplete visual data, such as depth cues and disrupted symmetry.

Our findings show that statistical techniques, such as feature importance metrics are essential for understanding the limitations of MLLMs. While GPT-4o handles simpler features effectively, its difficulty with complex stimuli highlights the need for incorporating human-like top-down inference mechanisms to improve robustness. Expanding this work with larger datasets and features like texture or shading can enhance our understanding of multi-modal AI systems. Statistical models can assist by identifying key limitations and guiding improvements for more robust performance.

\end{document}